\documentclass{ieeeaccess}
\usepackage{cite}
\usepackage{amsmath,amssymb,amsfonts}
\usepackage{algorithmic}
\usepackage{graphicx}
\usepackage{textcomp}
\graphicspath{{Figures/}}
\usepackage{caption}
\usepackage{subfig}
\usepackage{lipsum}
\usepackage{url}
\usepackage{amsthm}
\newtheorem{definition}{Definition}
\usepackage{balance}

\def\BibTeX{{\rm B\kern-.05em{\sc i\kern-.025em b}\kern-.08em
    T\kern-.1667em\lower.7ex\hbox{E}\kern-.125emX}}
\begin{document}
\history{Date of publication xxxx 00, 0000, date of current version xxxx 00, 0000.}
\doi{10.1109/xxxx}

\title{Identifying Stroke Indicators Using Rough Sets}
\author{
\uppercase{Muhammad Salman Pathan}\authorrefmark{1},
\uppercase{Jianbiao Zhang}\authorrefmark{1}
\IEEEmembership{Member, IEEE}, 
\uppercase{Deepu John} \authorrefmark{2}\IEEEmembership{Senior Member, IEEE},
\uppercase{Avishek Nag}\authorrefmark{2} \IEEEmembership{Senior Member, IEEE}, 
\uppercase{and Soumyabrata Dev}\authorrefmark{3 }\IEEEmembership{Member, IEEE}
}
\address[1]{Beijing Key Laboratory of Trusted Computing, Faculty of Information Technology, Beijing University of Technology, Beijing 100124, China}
\address[2]{University College Dublin, Dublin, Ireland}
\address[3]{ADAPT SFI Research Centre, Dublin, Ireland}

\tfootnote{The work in this paper has been supported by National Natural Science Foundation of China (No. 61971014). The  ADAPT  Centre  for  Digital  Content  Technology  is  funded under  the  SFI  Research  Centres Programme  (Grant  13/RC/2106)  and  is  co-funded  under  the  European  Regional  Development Fund.\\
With the spirit of reproducible research, the code to reproduce the simulations in this paper is shared at \url{https://github.com/Sammyy092/Identifying-stroke-indicators-using-rough-sets}.
}

\markboth
{Pathan \headeretal: Identifying Stroke Indicators Using Rough Sets}
{Pathan \headeretal: Identifying Stroke Indicators Using Rough Sets}

\corresp{Corresponding author: Soumyabrata Dev (e-mail: soumyabrata.dev@adaptcentre.ie).}

\begin{abstract}
Stroke is widely considered as the second most common cause of mortality. The adverse consequences of stroke have led to global interest and work for improving the management and diagnosis of stroke. Various techniques for data mining have been used globally for accurate prediction of occurrence of stroke based on the risk factors that are associated  with  the electronic health care records (EHRs) of the patients. In particular, EHRs routinely contain several thousands of features and most of them are redundant and irrelevant that need to be discarded to enhance the prediction accuracy. The choice of feature-selection methods can help in improving the prediction accuracy of the model and efficient data management of the archived input features.
In this paper, we systematically analyze the various features in EHR records for the detection of stroke. We propose a novel rough-set based technique for ranking the importance of the various EHR records in detecting stroke. Unlike the conventional rough-set techniques, our proposed technique can be applied on any dataset that comprises binary feature sets. We evaluated our proposed method in a publicly available dataset of EHR, and concluded that age, average glucose level, heart disease, and hypertension were the most essential attributes for detecting stroke in patients. Furthermore, we benchmarked the proposed technique with other popular feature-selection techniques. We obtained the best performance in ranking the importance of individual features in detecting stroke. 

\end{abstract}


\titlepgskip=-15pt

\maketitle

\section{Introduction}
\label{sec:introduction}

\PARstart
{S}troke is known as the second leading cause of death \cite{virani2020heart}. According to World Health Organization (WHO), the estimated figure of deaths leading from cardiovascular diseases rose to 17.7 million people in 2017, and around 6.7 million of them died due to stroke\cite{shikany2020abstract}. The prevalence and mortality of stroke are still on the rise \cite{mourguet2019increased}. World Stroke Day is observed every year where people are being educated about stroke and its prevention \cite{lindsay2019world}. A timely detection and prevention of stroke has become very essential to avoid its adverse consequences.

The field of medical sciences has observed tremendous improvements due to the rise in technological advancements over time. Most importantly, the Internet of Things (IoT) have made it easier to gather the data related to healthcare because of the availability of low-cost wearable devices \cite{gold2018clinical,ma2016big,mettler2016blockchain}. Tons of raw medical data is extracted from these devices to obtain knowledgeable patterns using various data-mining techniques. In addition, insights that are acquired are then used for decision making in healthcare sector and have proved as cost cutting element\cite{yadav2018mining}. Machine learning (ML) has established a potential space in the field of medical sciences in recent years. ML models can be applied on EHRs to predict the risk of having stroke on each patient efficiently\cite{chen2017disease}.

The predictive abilities of the models depend on the features/attributes selected from the data. The EHRs of the patients contain, collect, and archive several aspects of the patient's conditions. However, all the collected features in EHR may be useful for the detection of stroke. 
To improve the prediction performance of the machine learning models and reduce the machine training time, only important features associated with outcome should be selected for the prediction model \cite{amin2019identification}. Several techniques for data mining have also been proposed to extract the important and relevant features for predicting stroke occurrence \cite{esfahani2017cardiovascular,kolukisa2018evaluation,ratajczak2019automatic}. The major constraint of existing researches on risk assessment of stroke lies in the lack of proper systematic guidance for selection of features while building the prediction models, which is crucial for the model performance\cite{gupta2019apply}. Consequently, the most important research question of this study is how to select only the most discriminatory features from a very large dataset and understand the relative importance of each feature in predicting the stroke. We will only be able to detect the stroke efficiently, after the extraction of core risk factors that are highly associated with the onset of stroke.

The contribution of this work can be summarized as follows:
\begin{itemize}
    \item Identifying the important features in electronic health records for detecting stroke; 
   \item Proposal of a novel rough set methodology that can be used in binary feature data, in addition to the traditional feature data; and
    \item Release of open-source code on the proposed methodology and benchmarking methods to the research community. 
\end{itemize}

The rest of the article is organized as follows. The related work are explained in Section \ref{sec:related work}. In Section \ref{sec:prop}, the proposed methodology is described in detail. Section \ref{sec:results} discusses the results and benchmarks with previous approaches. Finally, Section \ref{sec:conc} concludes the paper.

\section{Related Work}
\label{sec:related work}
Due to the increasing number of patients of heart disease and high costs attached to it, researchers are working on solutions to prevent this disease, with an efficient treatment. Many proposals are published regarding the automatic detection of heart stroke using EHRs. However, the most important thing for accuracy in detecting stroke risk is to select and identify those features that influence the stroke outcome\cite{azhar2019comparative}. An identification of the most important features from a very high dimensional EHR is a pivotal issue from the medical point of view which can result in reducing the number of medical injuries.

\subsection{Feature Selection using Traditional Approaches}
In \cite{zhang2019stroke}, Zhang \textit{et al.} provides a study of an efficient feature selection method namely weighting-and ranking-based hybrid feature selection (WRHFS) to understand the probability of stroke. WRHFS employed a variety of filter-based feature selection models including information gain, standard deviation, and fisher score for weighing and ranking of the provided features. The research method selected 9 key features out of 28 features, on the basis of enough knowledge, for prediction of stroke. In another research carried by \cite{le2018automatic}, authors worked on a heart-disease-feature-selection method to extract important risk factors for heart disease prediction from a very large number of attributes. In this research, attributes are used based on their ranks. The ranks are given by Infinite Latent Feature Selection (ILFS) method which is is a feature selection algorithm based on probabilistic latent graph method. ILFS performs the ranking process by considering all the possible subsets of features exploiting the convergence properties of power series of matrices. The research used merely half of the total of 50 heart disease attributes and the result of the model was competitive.

It is obvious that the choice of an efficient feature selection method along with a robust classifier is extremely important for the heart disease classification problem. Models like support vector machine \cite{ayushi2019survey}, decision trees \cite{bin2017detection}, deep neural network \cite{panwar2019rehab}, and ensemble methods \cite{kavanagh2019evaluation} have achieved notable results to classify stroke outcomes when provided with relevant data. Authors of \cite{park2018bayesian} used important risk factors to present the evaluation of Bayesian networks for providing post-stroke results. Information gain method is used in this study for eliminating irrelevant features. The identified suitable feature set was provided as input to the Bayesian network. As per the results of the experiment, the features that were common for the prediction of post-stroke results included High-sensitivity C-reactive Protein (hsCPR), D-dimer, Initial NIH Stroke Scale (NIHSS), and Age. In another research  \cite{khosla2010integrated}, the authors presented a different algorithm for feature selection based on the conservative mean measure. The feature-selection algorithm was combined with the Support Vector Machine (SVM) classifier to classify the stroke outcome based on reduced feature set. A feature-selection model was presented in\cite{zhang2018risk} which combined SVM with the glow-worm swarm optimization algorithm on the basis of the standard deviation of the features. According to the outcomes, $6$ out of $28$ features were found to be most important for the risk associated with stroke that includes age, high blood pressure, Serum Creatinine (SCr), Lactate Dehydrogenase (LDH), and Alkaline Phosphatase (ALP) leaving behind the traditional factors including family history of stroke. A novel heart disease prediction system using ensemble deep learning and feature fusion method is presented in \cite{ali2020smart}. Firstly, a feature fusion approach is employed to generate rich heart disease dataset from raw data collected from different electronic sources. Next, information gain measure is used to eliminate irrelevant features. In addition, a conditional probability approach is utilized that computes a specific feature weight for each class in order to increase the prediction accuracy. Finally, an ensemble deep learning classifier is trained to predict heart disease in patients. From these works, it can be observed that feature selection methods can effectively increase the performance of prediction models in diagnosing the heart disease \cite{nourmohammadi2019new}.

\subsection{Feature Selection using Rough Set Theory}

Several factors including the choice of the input feature is an important factor in any machine learning task. An optimal subset of features that possess most of the information about the dataset can simplify the data description. Therefore, the method for selecting discriminant features from the medical dataset is the key issue for stroke diagnosis\cite{li2019medical}. Rough sets theory introduced by Pawlak\cite{pawlak1998rough}, is a new statistical approach to data mining and analysis that has been widely extended to many real-life applications in health care, manufacturing, finance, engineering, and others. Rough sets can obtain a subset of attributes that retains the discernible value of all the original attributes, by using the data without any additional information. Therefore it played a significant role in the selection of important features also called as feature selection or attribute reduction. One of the main advantages of rough sets is that it does not need any preliminary or additional information about data\cite{rissino2009rough}. Many techniques are developed in which the authors have used rough sets or has modified it to extract meaningful information from highly dimensional data sets \textit{i.e.,} influential features in order to provide better classification results  \cite{herbert2009criteria}. In the study conducted by \cite{rivas2018multi} authors tried to improve the feature selection using rough sets theory by proposing a genetic algorithm and rough-sets-based multi-granulation method which provided adequate results. The authors of\cite{liu2017hybrid} have worked on feature reduction and classification system which is based on the ReliefF and Rough Set (RFRS) method. ReliefF is a feature estimation algorithms which can accurately estimate the quality of features with strong dependencies. ReliefF assigns a weight to each feature based on filter approach. The feature estimation obtained from ReliefF is then used as the input for the rough set to reduce the feature set and select most important features for the classifier. To avoid the weakness of the classical rough sets when handling the large categorical datasets, the authors of \cite{wang2019fuzzzy} proposed a novel fuzzy rough set model to perform rough computations for categorical attributes. The experimental results show that proposed technique is effective in selecting meaning full attributes from large categorical datasets as compared to existing algorithms. In \cite{wang2019feature} the authors have proposed an improved the feature evaluation and selection strategy based on neighborhood rough sets. They proposed a new neighborhood self-information (NSI) technique which uses both the lower and upper approximations space of the target dataset for feature selection which is not the case in traditional rough set theory. Using the feature classification information by both upper and lower approximations, the proposed technique extracts most optimal feature subset from a large feature space.
Similarly, in \cite{fan2018attribute} a new rough set model named max-decision neighborhood rough set is developed for redundant feature reduction. In this work the dependence value criteria of rough set model is modified by enlarging the lower approximation set size. The proposed method was positively correlated with the classification accuracy of features during the experiments. In order to improve the fuzzy decisions for attribute reduction when applied to high-dimensional datasets, a new distance measure is introduced in fuzzy rough sets in \cite{wang2019fuzzy}. An iterative strategy for computing fuzzy rough dependency was adopted for feature evaluation where the parameter of distance measure is changed to obtained better results on very high-dimensional datasets. The proposed technique works well when applied to large datasets.

These related works conclude that rough sets can provide better results in terms of extracting meaningful features from a very high-dimensional dataset. However, rough sets are mostly suitable for datasets that possess numerical features. They cannot be directly applied to a dataset that comprises binary features. This is not an ideal case in the real world\footnote{The relevance value of the conditional attribute on the decision attribute becomes null, indicating no feature importance.}. Therefore, in this paper, we proposed a modified rough set technique that can be applied to datasets comprising both numerical and binary features. We have restricted the validation of our proposed rough-set methodology for the application of detecting stroke in electronic health records. However, our proposed novel rough set methodology can be easily translated to other applications.

\section{Proposed Rough Set Methodology}
\label{sec:prop}
The traditional rough set theory provides some mathematical methods to approximate the conventional sets in a set theory. By using rough sets, we can set the boundaries of the conventional sets in an approximate way. Rough set theory deals with imprecise, inconsistent, and incomplete information. Here we will first describe the rough sets and its related terminologies, and later define the important stroke-feature-selection method using rough sets.
\subsection{Rough Sets}

In rough sets, information is contained in a table that is often called as decision table. 
\begin{definition}
The decision table is defined as an approximation space ($U,L$), where $U={r_1,r_2,....r_n}$ is the finite non-empty set of observations also called as universe and $L$ is a family of attributes such that $L=C\cup D$ where $C$ and $D$ are condition and decision attributes. 
\end{definition}

For each observation in the decision table, we define a function $f:U \times L \rightarrow V$, where $f$ maps attribute $L$ to value domain of $V$. The function $f$ is also called as information function.

Tables can contain several objects which have the similar characteristics. A way to reduce the dimensionality of the table is to replace every set of object of having same properties with only one representative object. Such indiscernible objects are classified as reducts. With any reduct $P \in L$ there is an associated equivalence relation $IND(P)$.

\begin{definition}
The relation $IND(P)$ is called a  $P$-indiscernibility relation which is defined as: 
\[\mathrm{IND}(P)= \{(x, y)   \in   U^2 |   \forall m \in  P , f (x, m) =  f(y, m)\},\]
\end{definition}

Here $x$ and $y$ are two observations from the universe $U$ which are indiscernible from each other by attribute $P$, as both $x$ and $y$ are assigned similar value by function $f$. This partition of $U$ generated by $IND(P)$ is the family of all equivalence classes of $IND(P)$ and is denoted as:

\[U/IND(P) = f \{[x] | x \in U\}:\]

The equivalence classes of the $P$-indiscernibility relation are denoted by $[x]$.

Let us consider the following information table to explain in further details. Table~\ref{tab:table-equivclas} consists of two attributes: age and Lower Extremity Motor Score (LEMS) and seven observations. % as an example.

\begin{table}[htb]
\centering
\caption{Illustrative example of a information table comprising $7$ observations, and two conditional attributes (age and LEMS).}
\begin{tabular}{c|ccc}
\hline
Observations & Age & LEMS \\ \hline
$r_1$ & 16  & 50  \\
$r_2$ & 16  & 0  \\
$r_3$ & 31  & 1  \\
$r_4$ & 31  & 1  \\
$r_5$ & 46  & 26  \\
$r_6$ & 16  & 26  \\
$r_7$ & 46  & 26  \\ \hline
\end{tabular}
\label{tab:table-equivclas}
\end{table}

If we consider the complete set of the attributes $P=\{Age, LEMS\}$, we will obtain the following equivalence classes $\{r_1\}$, $\{r_2\}$, $\{r_3, r_4\}$, $\{r_5, r_7\}$, $\{r_6\}$ after applying indiscernibility function. As we can see that the rows in the third and fourth equivalence class \textit{i.e.,} $\{r_3, r_4\}$ and $\{r_5, r_7\}$ are not distinguishable from each other based on the available attributes. All the remaining rows are discernible from each other. It is obvious that different attributes lead to different results. If we consider only one attribute $P=\{age\}$, we will obtain the three different equivalence classes i.e. $\{r_1, r_2, r_6\}$, $\{r_3, r_4\}$ and $\{r_5, r_7\}$.

Suppose there is a conventional set $S$, which contains the set  of  observations  from U. Now the question is how to express the set $S$ using the information  in attribute set $P$? For this purpose, approximations (upper and lower) are generated by the roughset theory. The equivalence classes provide a solid foundation in order to construct the upper and lower approximations. 
\begin{definition}
The upper approximation and lower approximation are defined as:

\[\overline{P}(\mathit{S}) = \cup\left \{[x]|[x] \subseteq \mathit{S} \right \} ;\]

\[\underline{P}(\mathit{S}) = \cup\left \{[x]|[x] \cap \mathit{S} \neq \phi\right \} ;\]
\end{definition}

The lower approximation set $\underline{P}($S$)$ contains the observations that positively belong to $S$ \textit{i.e.,} the confirmed members, also referred as the positive region of the set $S$ denoted by $POS(S)$. Whereas the upper approximation set $\overline{P}(S)$ contains the observations which possibly belong to the conventional set $S$.

In order to access the quality of conditional attributes in determining the decision attributes, conventional rough-set-based techniques use an efficient criterion called relevance.

\begin{definition}
The relevance value is defined as the dependence of the conditional attribute $C$ on the decisional attribute $D$ and it is expressed as:

\[\gamma _C(D)=\frac{ \left | POS_c (D)\right |}{\left | U \right |}\]

\end{definition}

where $\left | . \right |$ denotes the cardinality of the set. We can recognize the most discriminatory attributes from the set $L$ by calculating this dependence value. The value of $\gamma _c$ varies from $0$ to $1$, where $0$ indicates independence and $1$ indicates that $D$ depends entirely on $C$.

\subsection{Proposed approach for binary features}
Rough set theory helps to represent ambiguous data at approximation levels and selects the most important features from a very large dataset. But there are some aspects that were not taken into consideration during the model development of conventional set-theory techniques. Pawlak’s model works well with nominal, non-binary data domains. However, real-world datasets also contain binary features. These binary features do not contain continuous data values but are binary in nature. As an illustration, the age attribute from the EHR records has continuous values across all the observations. On the other hand, attributes like gender are binary in nature as such attributes can possess \textit{select} values. In such cases of binary attributes, the relevance values of the conditional attributes on the decision attributes become null in most cases \textit{i.e.,} it is highly unlikely all the observations belonging to a single class of gender will either belong to either positive or negative samples of stroke. We attempt to solve this problem by generalizing the rough-set formulation into the universal case, wherein the conditional attributes can also possess binary data types. In this section we have presented the modified rough set model with an example to prove our claim.

Suppose that $U$ is a non-empty finite universe of observations for patients dataset $H_{i}$. We define $C_{i}$ as the set of conditional attributes from $A_{i}=\left \{ m \right \}$, where $m$ is the number of conditional attributes along with the vectorized decision attribute $D$ for $H_{i}$. We define a decision table $T_{i}$ as the set of rows and column where each row represents a medical observation and each column represents an attribute of patient.

Thus, for a dataset $H_{i}$, each value-attribute pair $\left ( q_{k},a_{j} \right )$ is assigned by a value $v_{i}^{kj}$ where $q_k$ is the k-th observation value of the patient and $a_j$ is the j-th attribute from the set $A_{i}$.

Our objective is to characterize the decision attribute $D$ from the knowledge of the reduct $P$ (\textit{cf.} Section~\ref{sec:prop}), a subset of conditional attribute set $C_{i}$. The objective is to identify the most discriminatory attributes which strongly depends on $D$. We define this dependence on a particular attribute by the proposed relevance-criterion-metric impact factor ($\mathcal{R}$) of the conditional attribute. Conditional attributes with a high impact factor value are better candidates for stroke detection.

\begin{definition}
We define our proposed relevance-criterion-metric impact factor ($\mathcal{R}$) for each conditional attribute C as:

\[\mathcal{R}=\left ( \frac{\mathcal{I}}{\left | U \right |} \right )\times 100\]

where $\mathcal{I}$ is the retention index. The retention index $\mathcal{I}$ is defined as the ratio of the number of elements of $C$ that are definite members of $D$. The retention index value can be obtained using following formula:

\[\mathcal{I}=\left ( \sum_{i=1}^{n} { \left | POS_c (D)\right |} \times \frac{ \left | POS_c (D)\right |}{\left | U_{j} \right |}\right )\]

Here $ POS_c (D)$ is the number of elements of $C$ that positively belongs to $D$ (cf.\ Section~\ref{sec:prop}). The $U_{j}$ is the total number of elements of $C$.
\end{definition}

In the following section, we have provided an illustrative example where the proposed methodology has been applied to obtain the dependence value for binary attributes available in a data set. 

\subsection{Illustrative Example}

Suppose we have a decision table having $7$ observations and $3$ conditional attributes as shown in Table~\ref{tab:table-example}. The attributes age and heart disease are conditional whereas stroke is decision attribute. In order to find the relevance-criterion-metric Impact Factor ($\mathcal{R}$) value of any $C$ for $D$, first we have to apply indiscernibility function for both the $C$ and $D$ attributes as described in Section~\ref{sec:prop}. 

\begin{table}[htb]
\centering
\caption{Illustrative example of a decision table comprising $7$ observations, and two conditional attributes (age and heart disease), and the decision attribute (stroke).}
\begin{tabular}{c|ccc}
\hline
Observations & Age & Heart Disease & Stroke\\ \hline
$x_1$ & 16  & 1 & 1 \\
$x_2$ & 16  & 0 & 0 \\
$x_3$ & 31  & 1 & 0 \\
$x_4$ & 31  & 1 & 1 \\
$x_5$ & 46  & 1 & 0 \\
$x_6$ & 16  & 0 & 1 \\
$x_7$ & 46  & 1 & 0 \\ \hline
\end{tabular}
\label{tab:table-example}
\end{table}

Suppose we want to find the Impact Factor ($\mathcal{R}$) value of age for stroke, the $IND (\mbox{age})$ would be $\{x_1, x_2, x_6\}$, $\{x_3, x_4\}$, and $\{x_5, x_7\}$. In the case of decision attribute, we are only interested in the row observations having the value ‘$1$’. We consider only the row observations that directly impact the positive occurrence of stroke. This is intuitive that the remaining elements have a negative label of stroke. This can be extended to multi-class classification problems, where we model our framework as multiple binary-classification tasks. Accordingly, $IND (\mbox{stroke})$ is  $\{x_1, x_4, x_6\}$, the set containing the observations having value $1$. We compute the reducts for both the columns/attributes, and obtain three sets for age and one set for stroke. 
The elements inside the sets are unique without any repetition.

\begin{figure}[htb]
    \centering
    \includegraphics[width=0.27\textwidth]{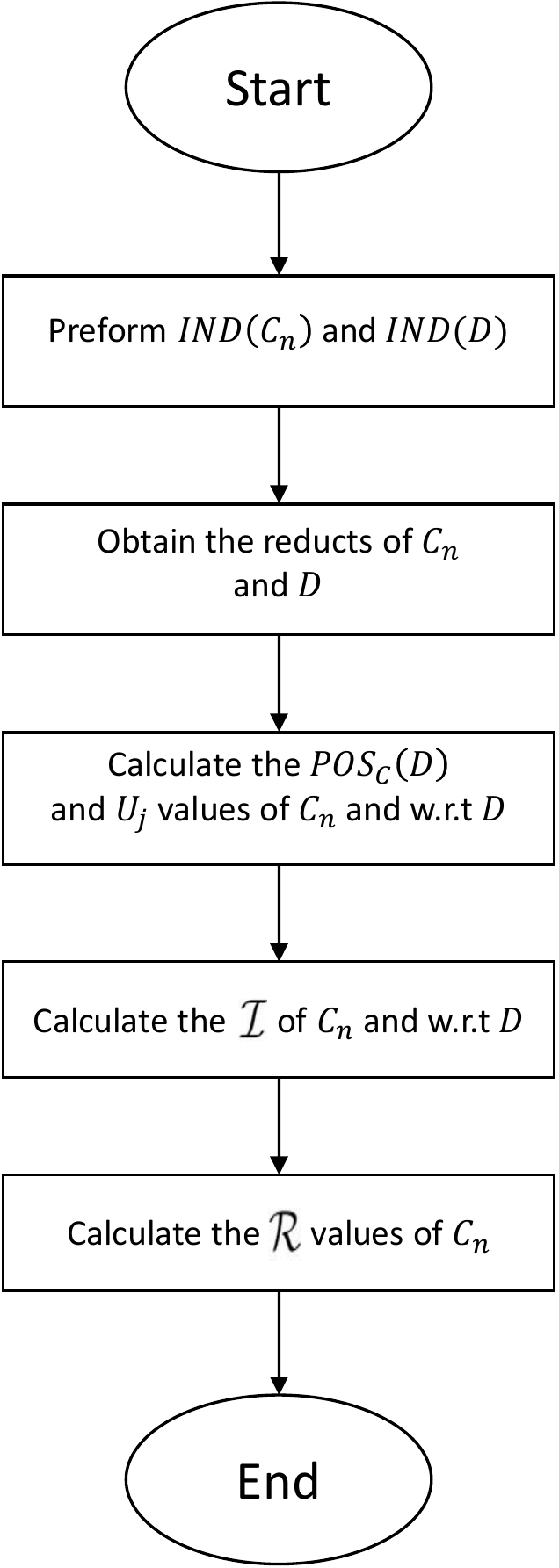}
    \caption{Flowchart of the proposed methodology describing each step for obtaining the $\mathcal{R}$ values.}
    \label{fig:flowchart}
\end{figure}

The next step in finding the $\mathcal{R}$ value is to compute the total retention-index value $\mathcal{I}$. As discussed before, $\mathcal{I}$ is the ratio of the elements from a conditional attribute which are definite member of decision attribute. For this purpose, we have to find the total number of the elements from attribute age which are included in attribute stroke. Comparing the first set of age with stroke, we found that only two elements \textit{i.e.}, $x_1$ and $x_6$ are present for the positive occurrence of stroke. So the $POS_c (D)$ value for the first set of age is $2$. The value of $U_{j}$ is $3$, \textit{i.e.,} the number of elements in the identified set. The value of $\mathcal{I}$ for the first set would be $1.33$. Accordingly, the total $\mathcal{I}$ value for the attribute age for stroke would be $1.66$. In this way the $\mathcal{R}$ value of attribute age would be $23.7$ where the $U$ is the total number of observations in the dataset \textit{i.e.} $7$. Applying similar steps for the heart disease column which is a binary attribute, the Impact Factor $\mathcal{R}$ value with respect to stroke is $18.28$. A detailed flowchart of the proposed methodology is shown in Figure~\ref{fig:flowchart}.

\section{Results \& Discussions}\label{sec:results}
\subsection{Dataset}
 
The Electronic Health Record (EHR) controlled by McKinsey \& Company was used as the dataset in our research which was a part of their healthcare hackathon\footnote{https://datahack.analyticsvidhya.com/contest/mckinsey-analytics-online-hackathon/}.The dataset is easily accessible as a free dataset repository\footnote{https://inclass.kaggle.com/asaumya/healthcare-dataset-stroke-data}. % which is a free dataset repository. 
The gathered data contained information of $29,072$ patients having $12$ common attributes. Out of the $12$ attributes, $11$ of them are input features including age, gender, marital status, patient identifier, work type, residence type (urban/rural), binary attribute heart disease condition,  
body mass index, smoking status of patient, glucose level and binary attribute hypertension indicating a patient is suffering from hypertension or not. The $12^{th}$ attribute is the binary output attribute indicating a patient is suffered stroke or not.

Continuous features are better understood when discretized into meaningful groups. Discretization is the method of converting continuous features into discrete features by creating a set of contiguous intervals spanning the range of feature values. In order to make the dataset more clear to understand, we have performed discretization on the $3$ attributes \textit{i.e.,} average glucose level, BMI, and age. Although data discretization is useful, we need to effectively pick the range of intervals/levels according to medical knowledge to obtain meaningful results. From the medical perspective, the glucose level is divided into four different levels \textit{i.e.} hypoglycemia, normoglycaemia, prediabetes, and diabetes\footnote{http://pharmwarthegame.blogspot.com/2018/11/blood-glucose-levels-chart.html}. Accordingly, we discretized the average glucose level attribute into $4$ levels. Similarly, the attribute BMI is discretized into $4$ levels, because the standard weight status categories associated with it are divided into $4$ levels \textit{i.e.} underweight, normal, overweight, and obese\footnote{https://www.news-medical.net/health/What-is-Body-Mass-Index-(BMI).aspx}. Evidences are accumulating that age have great impact on the distribution of stroke risk factors\cite{kelly2010influence}. The risk of stroke increases with the age, but it does not mean only elderly people have high risk of stroke. According to the medical research, strokes can happen at any age from infancy to adulthood\footnote{https://www.heartandstroke.ca/stroke/what-is-stroke/stroke-in-children}. The dataset contains patients of age ranging from $10$ to $80$ years old. The age attribute was discretized into $7$ different levels with the purpose of getting appropriate results for different age levels. In the following section, $10$ patient attributes were individually evaluated to identify its efficacy in the detection of stroke from electronic health records.

\subsection{Stroke Detection Performance}
In this section, we have evaluated the distinctive aspects of every input attribute for the classification task of the stroke. On the basis of previous analysis, we have used all of the $10$ attributes for classification task. The provided dataset is strongly unbalanced in nature. Only $548$ out of $29,072$ patients had positive occurrences of stroke, while the rest of $28,524$ records of patients showed no stroke condition. This unbalanced nature of the dataset lead to an issue in training any machine learning model. Thus, to control the adverse effects of the unbalanced data, we have employed a random down-sampling technique. We have made two classes where $548$ observations are referred as minority class and remaining $28,524$ observations are referred as majority class. A balanced dataset is created that consists of $548$ majority and $548$ minority observations. We choose all the $548$ observations from minority cases and $548$ random observations from total $28,524$ majority cases. Now the balanced dataset contains a total of $1096$ observations. $70\%$ of the balanced dataset was used for training and the remaining $30\%$ of the dataset was used for testing the model's performance.

\begin{figure}[htb]
    \centering
    \includegraphics[width=0.45\textwidth]{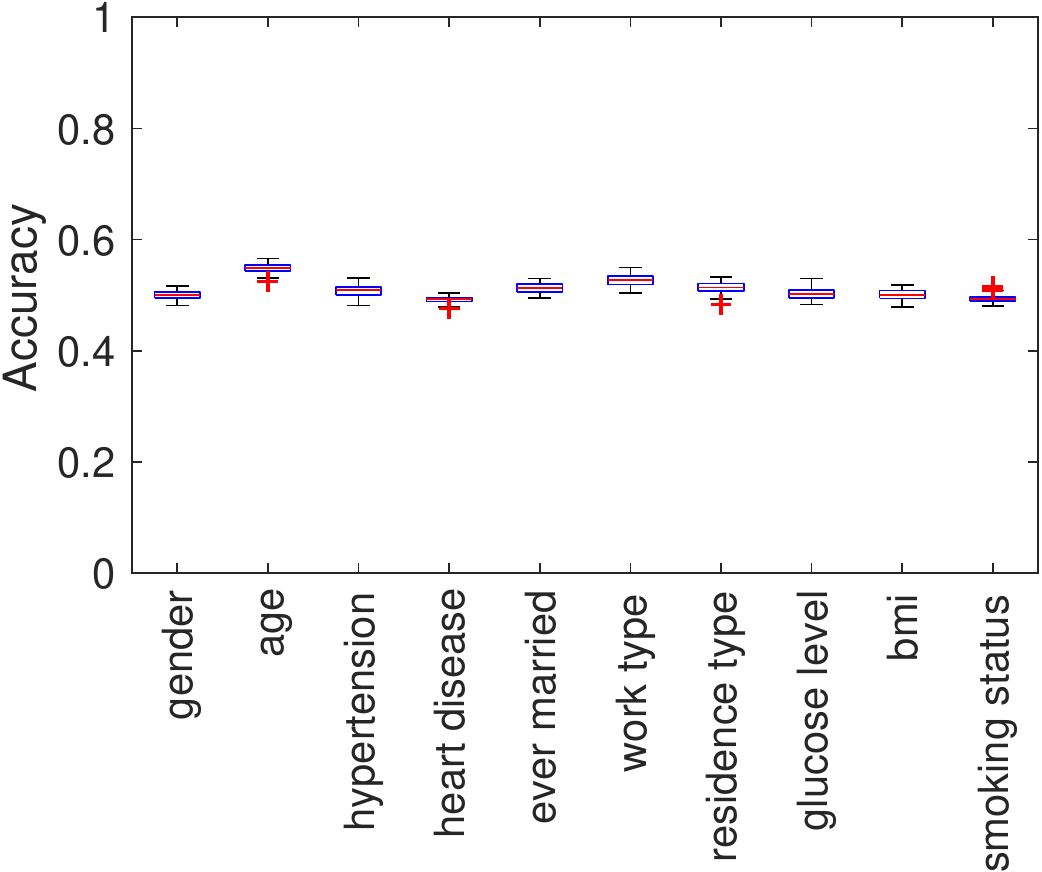}
    \caption{We show the stroke detection accuracy using different features of electronic health records. These results are obtained from random $100$ experiments.}
    \label{fig:boxplot}
\end{figure}

We have followed supervised approach of learning and trained SVM classifiers. We have used all the $10$ attributes separately as candidate feature vectors and trained $10$ different SVMs. Followed by this, the trained SVMs are used for stroke detection with the aim of checking the productiveness of the respective attributes in the classification task. For the elimination of the sampling bias, we performed $100$ random down-sampling experiments. The classification results for all the $10$ attributes are shown in Figure~\ref{fig:boxplot}. We observed that feature age have comparatively higher accurate results as reported in Figure~\ref{fig:boxplot}. However, every feature does have specific discriminative power to identify the stroke.

In order to perform an objective evaluation of the various attributes in the dataset, we have computed the precision, recall, F-score, and accuracy. % as the evaluation matrices. 
In the binary classification task, $TP$, $FP$, $TN$, and $FN$ are denoted as the true positives, false positives, true negatives, and false negatives respectively. Accordingly, the precision is defined as the ratio of correctly predicted observations to the total predicted observations. %, calculated as $TP/(TP+FP)$. 
The recall is the ratio of correctly predicted observations to the all observations in the identified class.  %$TP/(TP+FN)$. 
The F-Score is defined as the harmonic mean of Precision and Recall. Finally, the accuracy is defined as the ratio of correctly predicted observations to the total number of observations. They are computed as:

\[\mbox{Precision} =\frac{TP}{(TP + FP)}\]

\[\mbox{Recall} =\frac{TP}{(TP + FN)}\]

\[\mbox{F-score} =\frac{2\times\mbox{Precision}\times\mbox{Recall}}{(\mbox{Precision}+\mbox{Recall})}\]

\[\mbox{Accuracy} =\frac{(TP + TN)}{(TP + TN + FP + FN)}\].

We compute the precision, recall, F-score, and accuracy values of all the different features in the detecting stroke. Table~\ref{tab:score} summarizes the results. Each value is the average value computed across $100$ random sampling experiments. We observe that most of the features have similar performance across all the different metrics. However, the age feature performs slightly better as compared to the other features, as can also be seen from Fig.~\ref{fig:boxplot}. 

\begin{table}[htb]
\centering
\caption{Performance evaluation of the different attributes for the detection of stroke from the electronic health records. Each values are computed from the average value obtained from $100$ random sampling experiments on the balanced dataset.}
\begin{tabular}{r|cccc}
\hline
\textbf{Feature} & \textbf{Precision} & \textbf{Recall} & \textbf{F-Score} & \textbf{Accuracy}\\ \hline
gender & 0.50  & 0.32 & 0.30 & 0.50\\
age & 0.56  & 0.43 & 0.49 & 0.55\\
hypertension &  0.51  & 0.52 & 0.49 & 0.51\\
heart disease & 0.49  & 0.50 & 0.37 & 0.49\\
ever married & 0.63  & 0.06 & 0.11 & 0.51\\
work type & 0.52  & 0.76 & 0.62 & 0.53\\
residence type & 0.51  & 0.84 & 0.63 & 0.51\\
glucose level & 0.51  & 0.51 & 0.48 & 0.50\\
bmi & 0.52  & 0.38 & 0.34 & 0.50\\
smoking status & 0.49  & 0.39 & 0.32 & 0.49\\ \hline
\end{tabular}
\label{tab:score}
\end{table}

\subsection{Impact Factor scores}
In this section, we measure the dependence of all the attributes and report their corresponding proposed Impact Factor ($\mathcal{R}$) values using our proposed rough-set-based method. We observe that certain attributes such as age, hypertension, average glucose level, and heart disease have higher $\mathcal{R}$ values. As shown in Table~\ref{tab:rel-val}, the $\mathcal{R}$ scores for age, hypertension, average glucose level, and heart disease are $0.0921$, $0.0469$, $0.0476$, and $0.0559$ making these attributes favorable candidates for stroke prediction as compared to other attributes. On the other hand, attributes like gender, residence type, smoking status, and BMI have low $\mathcal{R}$ scores, depicting its low relevance to the decision attribute. Therefore, the latter attributes are not conducive for stroke prediction. 
\begin{table}[htb]
\small  
\centering
\caption{Average impact factor value across all the features of electronic health records. The most relevant factors that contributes to stroke are highlighted in bold.}
\begin{tabular}{ p{18mm} | p{9mm} || p{20mm} | p{9mm}}
\hline
gender & 0.0358 & age & \textbf{0.0921} \\ 
hypertension & \textbf{0.0469} & heart disease & \textbf{0.0559} \\
ever married & 0.0397 & work type & 0.0405\\
residence type & 0.0355 & average glucose level & \textbf{0.0476}\\
BMI & 0.0358 & smoking status & 0.0369\\
\hline
\end{tabular}
\label{tab:rel-val}
\end{table}

\begin{figure*}
\centering
\subfloat[Bimodality]{\includegraphics[height=0.35\textwidth]{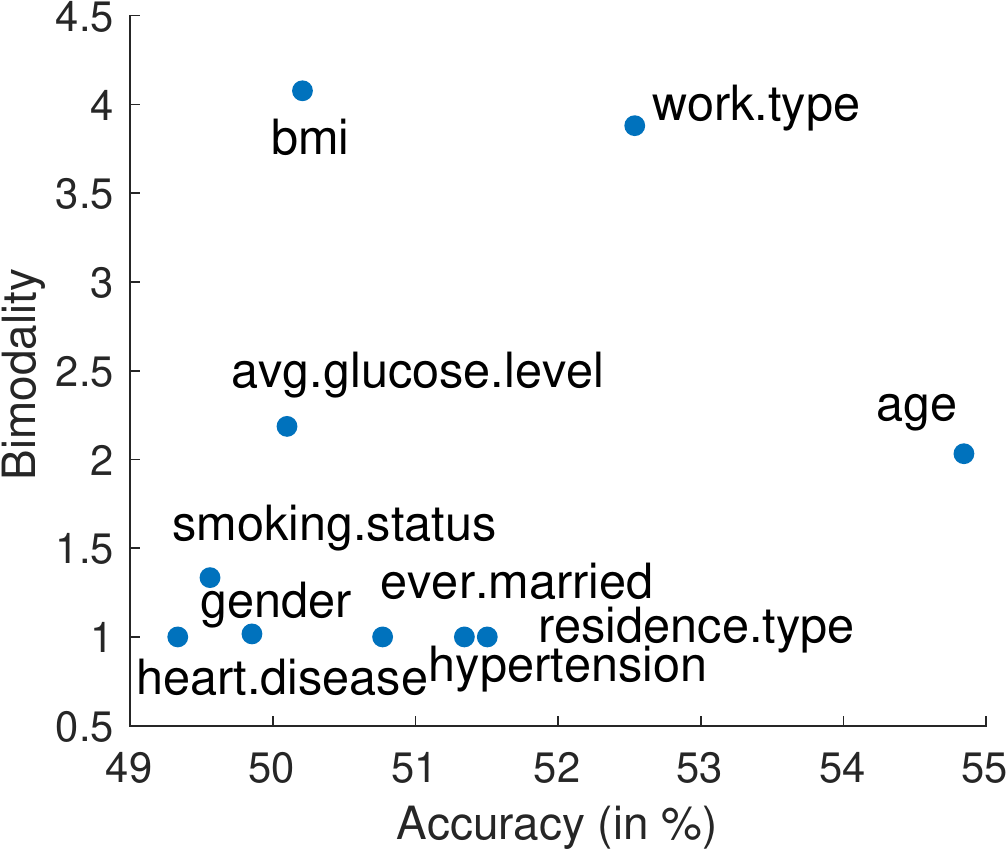}}
\subfloat[Loading factor]{\includegraphics[height=0.35\textwidth]{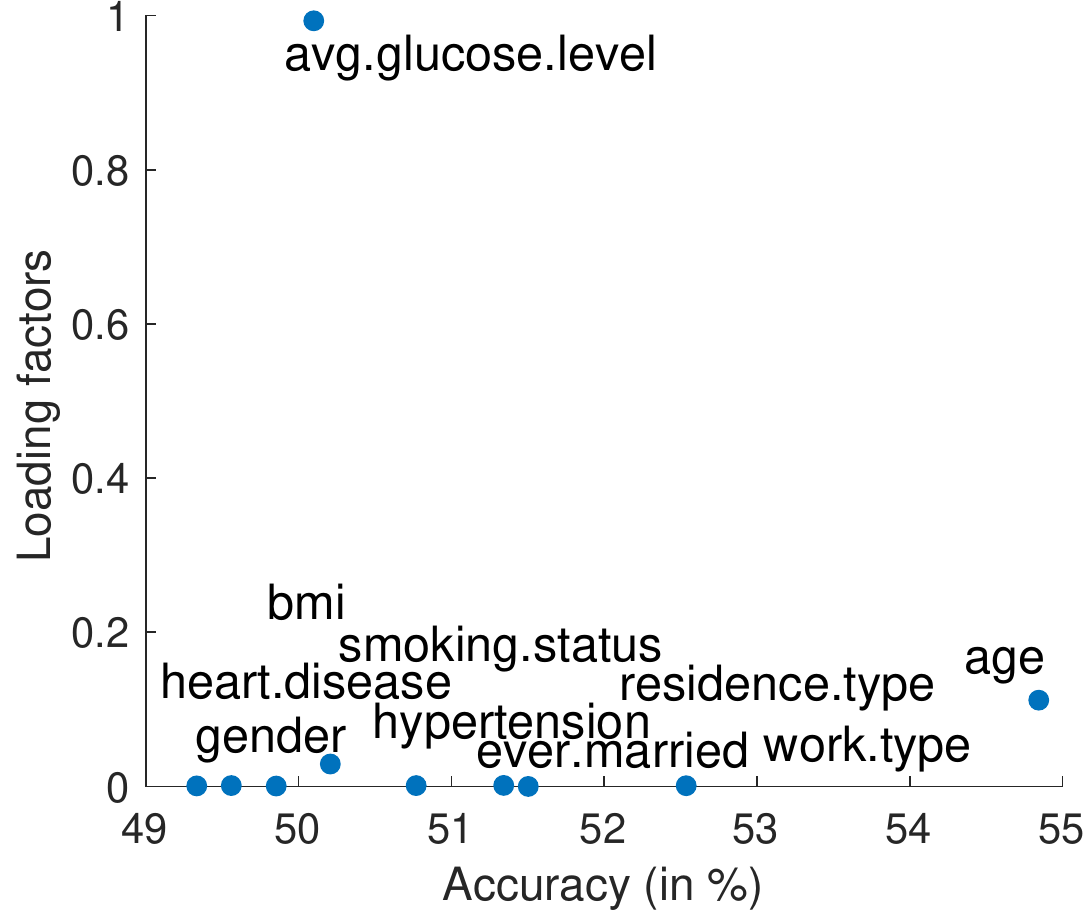}}\\
\subfloat[Wilcoxon]{\includegraphics[height=0.35\textwidth]{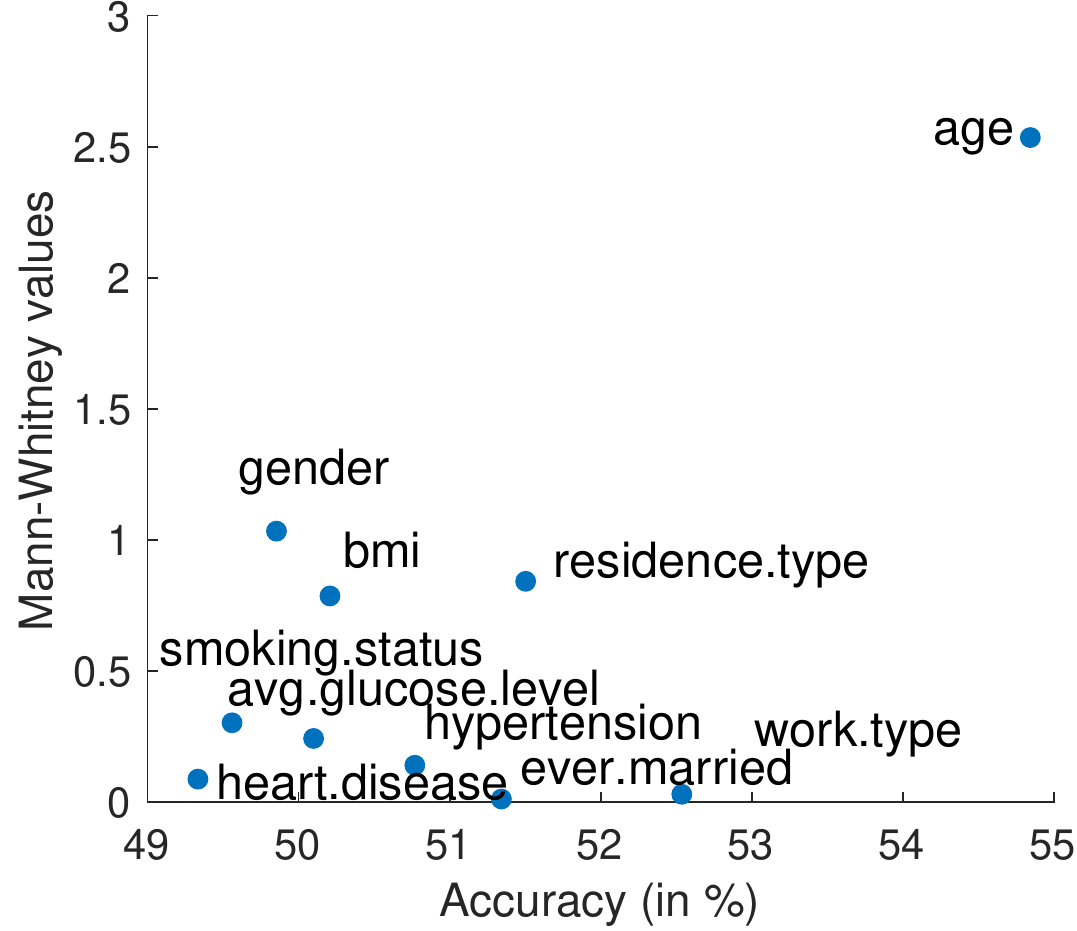}}
\subfloat[Proposed approach]{\includegraphics[height=0.35\textwidth]{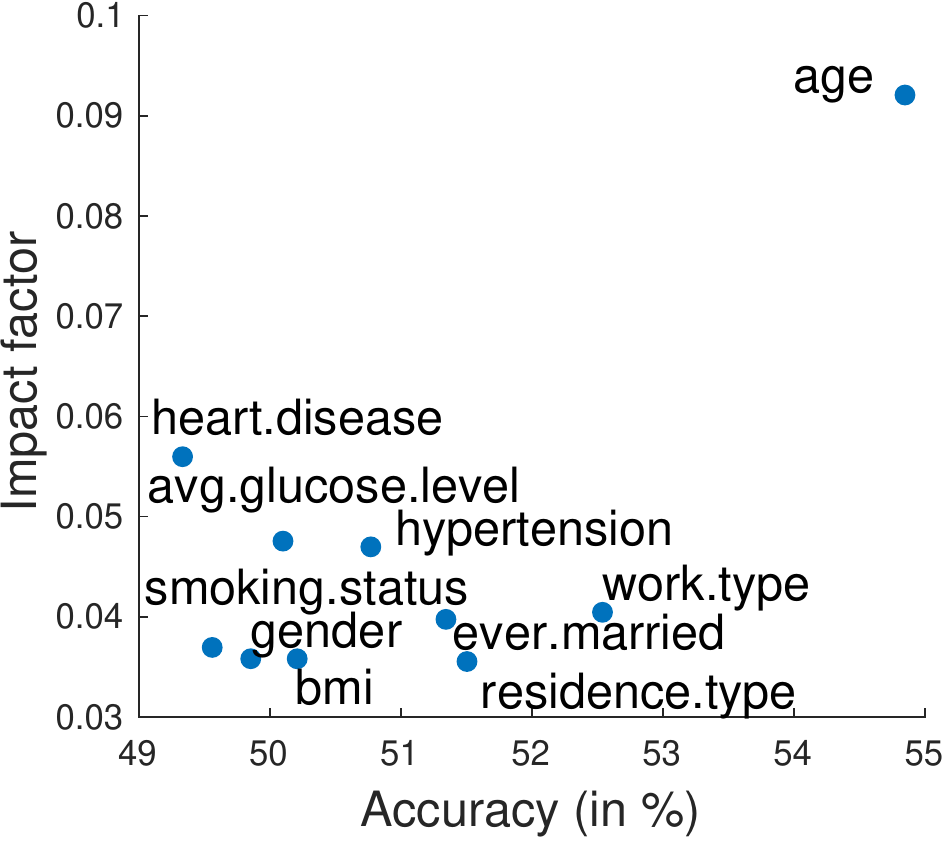}}
\caption{Scatter plot between average accuracy (ground-truth) values and (a) bimodality, (b) loading factor, (c) wilcoxon, and (d) proposed relevance value.}
\label{fig:scatter-result}
\end{figure*}

\subsection{Benchmarking}

In this section, we benchmark our proposed rough-set-based method with other models. Those features are favorable that exhibit higher bimodal behavior. Thus, we have extracted Pearson’s Bimodality Index (PBI) \cite{knapp2007bimodality} for all the $10$ attributes. Those PBI values that are closer to unity depict highest bimodality, whereas the values that are greater than $1$ depict lowest bimodality. We have also measured the loading factors values. Factor loadings can vary from $-1$ to $1$ having a similar characteristic to that of correlation coefficients. A loading factor of $0$ indicates no effect. 
At the end, the Wilcoxon test, which is a non-parametric statistical test that compares two paired groups, was done. This test calculates the actual difference between the set of pairs and provides analysis of the differences.

We have checked the correlation between the average accuracy scores of each attributes and the normalized scores attained from different models. Table~\ref{tab:corr-table} displays the correlation results of stroke classification accuracy with ranking scores obtained using different models. Our proposed rough set based method has the highest correlation value as compared with others.

\begin{table}[htb]
\centering
\caption{Correlation of stroke classification accuracy with ranking scores obtained using different methods. The best performance is indicated in bold.}
\begin{tabular}{ll}
\hline
\textbf{Methods}           & \textbf{Correlation (r)} \\ \hline
Proposed approach & \textbf{0.675}  \\
Bimodality        & 0.239        \\
Loading factors   & -0.106          \\
Wilcoxon          & 0.648          \\ \hline
\end{tabular}
\label{tab:corr-table}
\end{table}

Figure~\ref{fig:scatter-result} shows the respective scatter plots. Our proposed method achieves the highest correlation coefficient \textit{i.e.} $r = 0.675$. From Fig.~\ref{fig:scatter-result}(d), we can clearly see that attributes age, heart disease, average glucose level, and hypertension are better candidates for stroke prediction. Similarly, attributes like gender, residence type, and BMI have lower relevance scores and are poor candidates. The impact factor scores obtained from the various attributes are highly correlated with the accuracy percentage obtained from the individual attributes. We obtain the highest correlation value of $0.675$, as compared to the other benchmarking methods. Hence from the Fig.~\ref{fig:scatter-result} we came to the conclusion that our proposed feature-selection technique based on rough sets theory is useful for ranking important attributes and identifying those that are most suitable ones for the prediction of stroke as compared to the other techniques.

\subsection{Performance analysis due to size of dataset}
\begin{figure}[htb]
    \centering
    \includegraphics[width=0.45\textwidth]{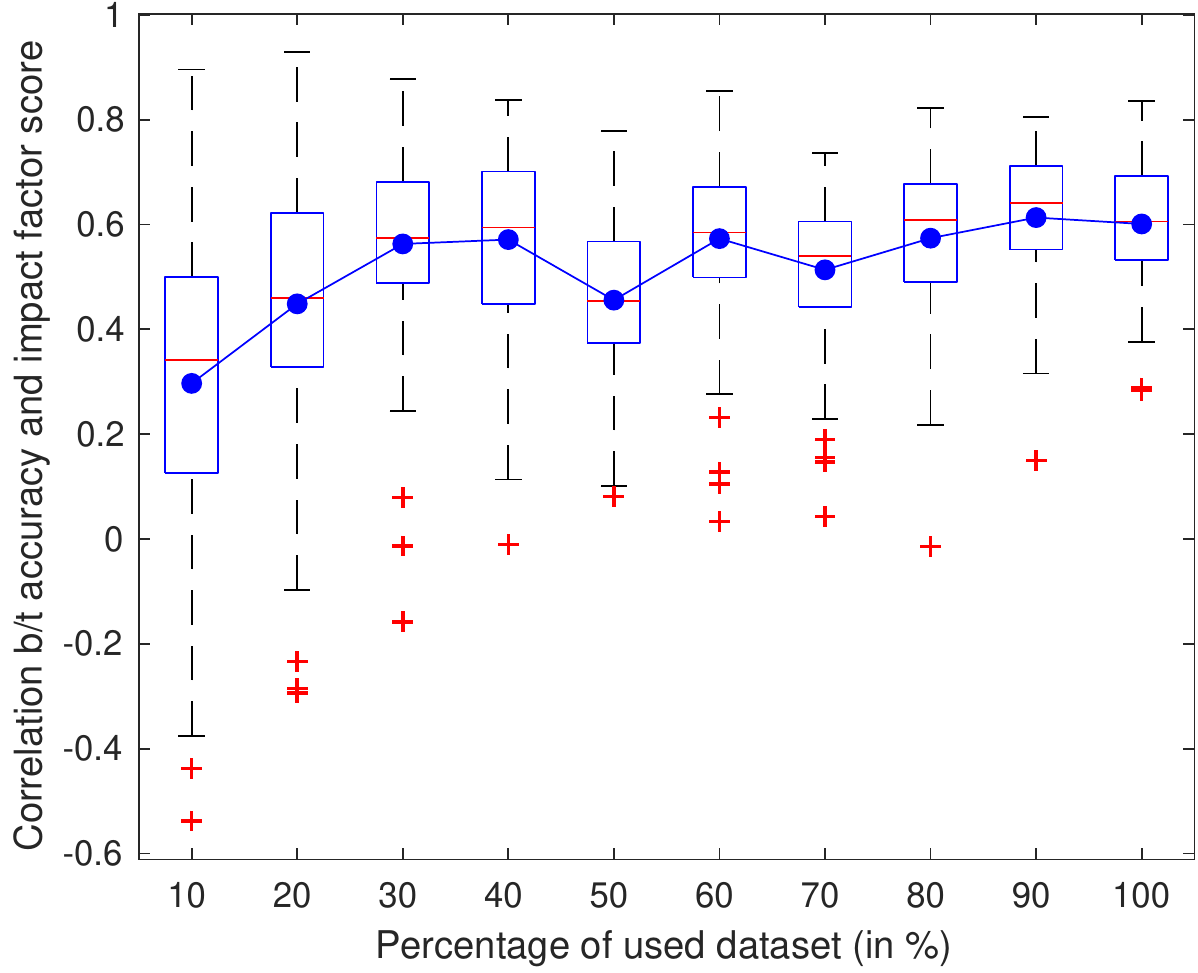}
    \caption{We plot the relationship of the correlation between accuracy and impact factor scores with respect to the percentage of used dataset. The box plots show the results obtained from $100$ experiments. The \textit{blue} solid line indicates the mean correlation values across the different percentage.}
    \label{fig:size-impact}
\end{figure}

It is obvious that the amount of data in a database can increase or decrease the prediction performance. A dataset having enough information in it can help a prediction model to generate accurate results. In this section we have accessed the performance of our proposed method by varying the size of data set. We have divided the data set into 10 fractions ranging from $10\%$ to $100\%$. We took the random samples of data from the original database. 
As we can see from the Fig.~\ref{fig:size-impact}, there is not significant variance in the correlation values. We can see that when the data is $10\%$ and $40\%$ the correlation values are low as compared to other fraction and the relevance value is high when the database fraction is at $70\%$ and $100\%$. According to the relevance value scores, we can observe that the amount of data and the knowledge inside the database can heavily affect the result outcomes.

\section{Conclusion}\label{sec:conc}
Efficient stroke-detection methods have been increasingly studied in recent years. The selection of important features from the high dimensional medical dataset is critical to the prediction model’s performance. Existing research on automatic detection of stroke risk through data mining techniques faces a significant challenge in the selection of effective features as predictive cues. RST is an optimized feature selection method which helps in identifying and ranking the most important features in a large dataset. Its main idea is to approximately describe inaccurate or indefinite knowledge with known knowledge. In this research, we have proposed an efficient feature selection technique based on rough set theory. The proposed technique can identify the critical stroke indicators from a very large dataset for building effective prediction models for stroke detection. Furthermore, we modified the rough sets technique so that it can be applied on medical datasets having binary feature sets referring to real-world applications which were not feasible using traditional rough sets technique. We applied the proposed technique on a dataset containing $29,072$ patient records and having $10$ common attributes. We found $4$ most important attributes \textit{i.e.,} age, heart disease, hypertension, and average glucose level which contributes the maximum in the detection of stroke as compared to other attributes. We also benchmarked the proposed technique with other popular feature-selection techniques. The obtained correlation values highlighted that our proposed technique was positively correlational with the classification accuracy and is better in identifying and ranking the most important attributes in the dataset for the prediction of stroke. Finally, we observed the proposed technique on different fractions of the dataset. According to the results we found that the size of the database can influence the result outcome. Our future works include the proposal of a machine-learning framework using the identified features in electronic health records for an improved stroke detection. We also plan to investigate the use of our proposed rough-set technique in other related applications.
\appendices

\bibliographystyle{IEEEtran.bst}


\begin{IEEEbiography}[{\includegraphics[width=1in,height=1.25in,clip,keepaspectratio]{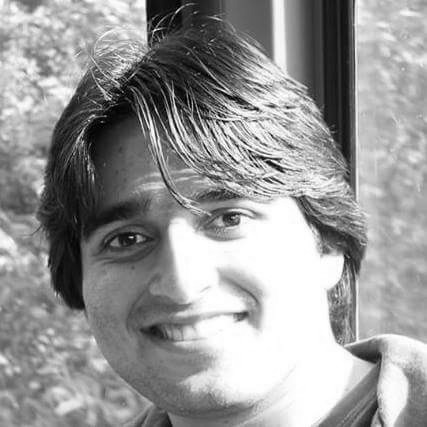}}]{Muhammad Salman Pathan} received his PhD degree from Beijing University of Technology (BJUT), China. Currently, he is working as Post-Doctoral fellow at the Faculty of Information Technology, BJUT, China. His research interest are machine Learning, information security, wireless networks. He has published several research papers in his area.
\end{IEEEbiography}

\begin{IEEEbiography}[{\includegraphics[width=1in,height=1.25in,clip,keepaspectratio]{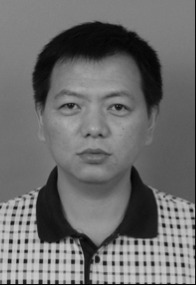}}]{Jianbiao Zhang} received his B.S., M.S. and Ph.D. degree in computer science, all from Northwestern Polytechnic University in 1992, 1995 and 1999, Xi’an, China. From 1999 to 2001, he was a postdoctoral fellow in Beihang University, Beijing, China. Now, he is a professor and Ph.D. supervisor in Faculty of Information Technology, Beijing University of Technology. His research interests include network and information security, trusted computing. He has published over 80 journal/conference papers.
\end{IEEEbiography}

\begin{IEEEbiography}[{\includegraphics[width=1in,height=1.25in,clip,keepaspectratio]{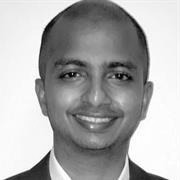}}]{Deepu John} received the B.Tech degree in Electronics and Communication engineering from University of Kerala, India, in 2002 and the MSc and PhD degrees in electrical engineering from National University Singapore, Singapore in 2008 and 2014 respectively. He is a recipient of Institution of Engineers Singapore Prestigious Engineering Achievement Award (2011), Best design award at Asian Solid State Circuit Conference (2013), IEEE Young Professionals, Region 10 individual award (2013). He served as a member of technical program committee for IEEE conferences TENCON 2009, ASICON 2015, TENCON 2016. He is a reviewer of several IEEE journals and conferences and serves as an Associate Editor for IEEE Transactions on Biomedical Circuits and Systems. His research interests include low power biomedical circuit design, energy efficient signal processing and wearable healthcare devices.
\end{IEEEbiography}

\begin{IEEEbiography}[{\includegraphics[width=1in,height=1.25in,clip,keepaspectratio]{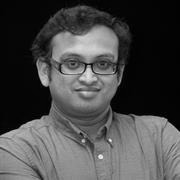}}]{Avishek Nag} is currently an Assistant Professor in the School of Electrical and Electronic Engineering at University College Dublin in Ireland. Dr Nag received the BE (Honours) degree from Jadavpur University, Kolkata, India, in 2005, the MTech degree from the Indian Institute of Technology, Kharagpur, India, in 2007, and the PhD degree from University of California, Davis in 2012. He worked as a research associate at the CONNECT centre for future networks and communication in Trinity College Dublin, before joining University College Dublin. His research interests include, but are not limited to Cross-layer optimisation in Wired and Wireless Networks, Network Reliability, Mathematics of Networks (Optimisation, Graph Theory), Network Virtualisation, Software-Defined Networks, Machine Learning, Data Analytics, Blockchain, and the Internet of Things. Dr Nag is a senior member of the Institute of electronics and electrical engineers (IEEE) and also the outreach lead for Ireland for the IEEE UK and Ireland Blockchain Group.
\end{IEEEbiography}

\begin{IEEEbiography}[{\includegraphics[width=1in,height=1.25in,clip,keepaspectratio]{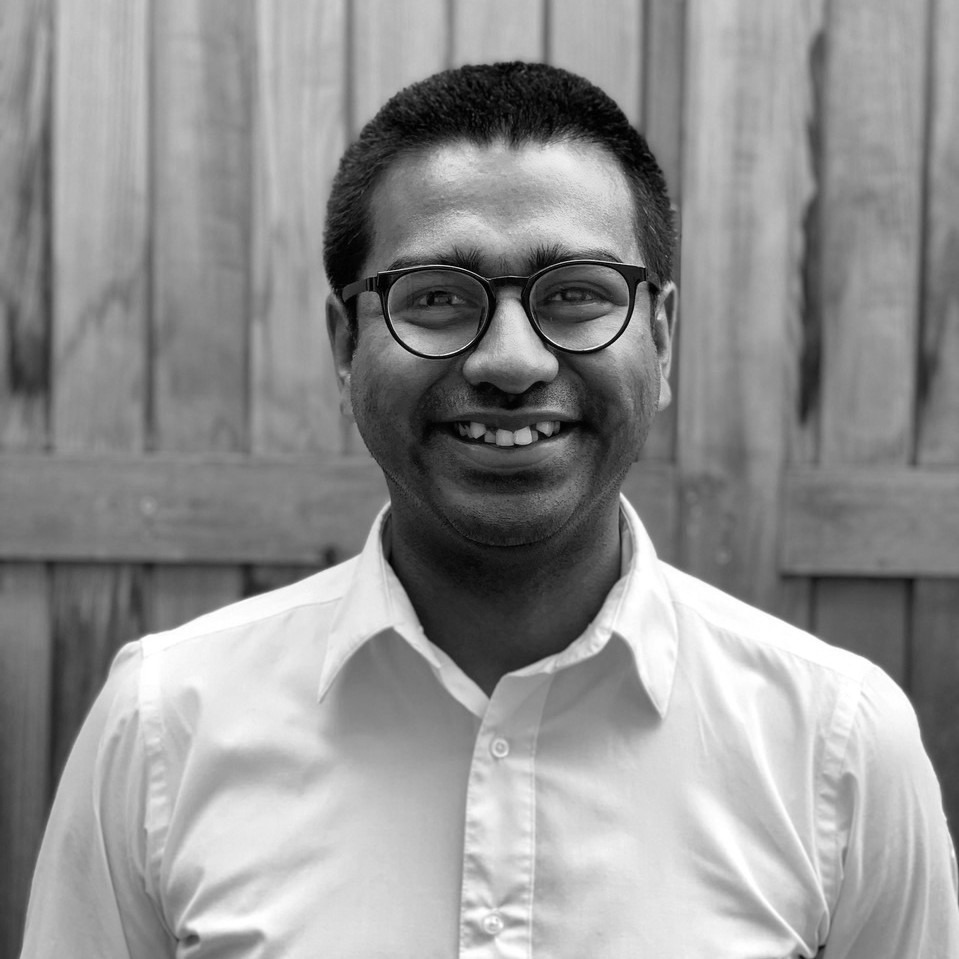}}]{Soumyabrata Dev} (S'09-M'17) is an Assistant Professor in the School of Computer Science at University College Dublin. He obtained his PhD from Nanyang Technological University (NTU) Singapore in 2017. From August-December 2015, he was a visiting doctoral student at Audiovisual Communication Laboratory (LCAV), \'{E}cole Polytechnique F\'{e}d\'{e}rale de Lausanne (EPFL), Switzerland. He worked at Ericsson as a network engineer from 2010 to 2012. Prior to this, he graduated \emph{summa cum laude} from National Institute of Technology Silchar, India with a BTech in 2010. His research interests include  remote sensing, statistical image processing, machine learning, and deep learning.
\end{IEEEbiography}

\balance 

\EOD

\end{document}